\documentclass[10pt,journal,compsoc]{IEEEtran}
\ifCLASSOPTIONcompsoc
\else
\fi
\ifCLASSINFOpdf
\else
\fi
\usepackage[cmex10]{amsmath}
\usepackage{algorithmic}
\usepackage{algorithm}

\usepackage{array}
\newcommand{\PreserveBackslash}[1]
{\let\temp=\\#1\let\\=\temp}
\newcolumntype{C}[1]{>{\PreserveBackslash\centering}p{#1}}
\newcolumntype{L}[1]{>{\PreserveBackslash\raggedright}p{#1}}

\usepackage{graphicx}
\usepackage{caption}
\usepackage{enumitem}
\usepackage{url}
\usepackage{xcolor}
\definecolor{newcolor}{rgb}{.8,.349,.1}
\usepackage{framed,multirow}
\usepackage{epstopdf}
\usepackage{amsmath}
\usepackage{algorithm}
\usepackage{algorithmic}
\usepackage{bm}
\usepackage{amssymb}
\usepackage{latexsym}

\def\eg{{e.g.,\ }}
\def\ie{{i.e.,\ }}
\def\etal{{et al.\ }}

\def\y{{\mathbf{y}}}
\def\z{{\mathbf{z}}}
\def\n{{\mathbf{n}}}
\def\u{{\mathbf{u}}}
\def\v{{\mathbf{v}}}
\def\R{{\mathbf{R}}}
\def\S{{\mathbf{S}}}
\def\I{{\mathbf{I}}}
\def\L{{\mathbf{L}}}
\def\N{{\mathbf{N}}}

\begin{document}
%
\title{Color naming guided intrinsic image decomposition}
%
%
%
%

\author{Yuanliu~Liu, 
        ~Zejian~Yuan$^*$,~\IEEEmembership{Member,~IEEE}, 
\IEEEcompsocitemizethanks{\IEEEcompsocthanksitem Y. Liu and Z. Yuan are with the Institute of Artificial Intelligence and Robotics, Xi'an Jiaotong University, Xi'an,
Shaanxi, 710049 China. Z. Yuan is the corresponding author.\protect\\
E-mail: liuyuanliu88@gmail.com, yuan.ze.jian@mail.xjtu.edu.cn}}

\IEEEtitleabstractindextext{%
\begin{abstract}
Intrinsic image decomposition is a severely under-constrained problem. User interactions can help to reduce the ambiguity of the decomposition considerably. The traditional way of user interaction is to draw scribbles that indicate regions with constant reflectance or shading. However the effect scopes of the scribbles are quite limited, so dozens of scribbles are often needed to rectify the whole decomposition, which is time consuming. In this paper we propose an efficient way of user interaction that users need only to annotate the color composition of the image. Color composition reveals the global distribution of reflectance, so it can help to adapt the whole decomposition directly. We build a generative model of the process that the albedo of the material produces both the reflectance through imaging and the color labels by color naming. Our model fuses effectively the physical properties of image formation and the top-down information from human color perception. Experimental results show that color naming can improve the performance of intrinsic image decomposition, especially in cleaning the shadows left in reflectance and solving the color constancy problem.
\end{abstract}

}

\maketitle

\IEEEdisplaynontitleabstractindextext

\IEEEpeerreviewmaketitle

\section{Introduction}
\label{sec:introduction}
%
%
%
%
Intrinsic image decomposition is proposed by \cite{Barrow_CVR78}. They represent each physical characteristic of the scene by a separate image, and all these intrinsic images are recovered from a single image. The primarily addressed intrinsic images are shading, reflectance, distance, and orientation. Over time, the types of intrinsic images have been limited to reflectance and shading \cite{intrinsic_dataset,Horn1974277,Tappen_05}. The reflectance is the albedo of the material, while the shading records the illumination that strikes the surface. The observed image is basically the pointwise product of the reflectance and the shading.

Intrinsic image decomposition is an important preprocess for many computer graphics and computer vision problems. One important application is image editing. After separating the factors of the scene, users can edit one or more factor while keeping the others untouched. More specifically, recoloring \cite{Beigpour_ICCV11,CarrollRA11}, retexturing \cite{Yan10Re-texuring}, colorization \cite{IntrinsicColorization}, and makeup simulation \cite{Makeup} change the reflectance, and relighting replaces the illuminants \cite{AppGen,Relighting}. In terms of computer vision, using the reflectance instead of the raw image for object recognition can increase the robustness to illumination \cite{IID_Obj_Attr}. Reflectance is also used for face alignment \cite{FaceAlignment}. The shading is widely used to recover the 3D shape of the objects by shape-from-shading techniques \cite{Shape_from_shading_origin,Zhang99}. 

Despite the potential of being a valuable preprocess, intrinsic image decomposition itself is a challenging task that limits its applicability. In essence, intrinsic image decomposition is an underconstrained problem. There are infinite combinations of reflectance and shading that can reproduce the input image. An intuitive solution is to incorporate more constraints to reduce the ambiguity. Usually, the constraints are raised from the prior knowledge of general scenes. The most widely used priors are the local smoothness of shading \cite{Chen_ICCV13,gehler11nips,Lee_ECCV12,ShenCVPR08}, the piecewise constancy of reflectance \cite{BarronECCV12,Bi2015L1Intrinsic,Chen_ICCV13,ShenCVPR08,Shen_PAMI13}, and the global sparsity of reflectance \cite{BarronECCV12,gehler11nips,Garces2012,Nie_2014,Shen_PAMI13}. However, these general priors are not applicable to irregular images or regions. For example, the shading is not smooth at shadow edges. More specific constraints come from the statistics of training datasets \cite{BarronPAMI15,MaloneyJOSA86}. The illuminants and reflectance are supposed to locate at a restricted area or certain peaks in the color space, so the solution space of intrinsic image decomposition will be significantly reduced. These data-dependent constraints can hardly be generalized to other datasets with different statistics.


When reliable prior information is unavailable, we have to seek help from user interactions.
The human visual system is able to perceive the intrinsic reflectance of the surfaces by the mechanism of \textbf{color constancy} \cite{Color_vision}. Advanced intelligence of discernment, inference, and recognition from the sensation of lights also plays an important role, as pointed out by Alhazen about a thousand years ago. These abilities enable us to understand the scene thoroughly, so we can infer the intrinsic images in a top-down fashion.

\cite{BousseauTOG2009} proposed the first user-assisted method of intrinsic image decomposition . They draw scribbles on the image to tell which areas have constant reflectance or shading. These scribbles help solving the ambiguities in determining whether the difference of luminance is caused by reflectance change or shading variation. People are good at finding out those planar areas whose shading change little. They can also locate the shading or shadow edges on surfaces of uniform reflectance according to the geometric or color information. To our best knowledge, all the user-assisted methods of intrinsic image/video decomposition follow this way \cite{Bonneel2014IntrinsicVideo,CarrollRA11,AppGen,OptScribbles,Ye2014IntrinsicVideo,Zhao_PAMI13}. The problem is that the constancy of reflectance or shading often appears in very local regions, so quite a lot of scribbles are needed to cover enough portion of the image. It is quite time-consuming to draw dozens of scribbles carefully on narrow regions. In addition, there is no scribble suitable for regions where both the shading and the reflectance are changing.

\begin{figure}
\begin{center}
\includegraphics[width=1.0\linewidth]{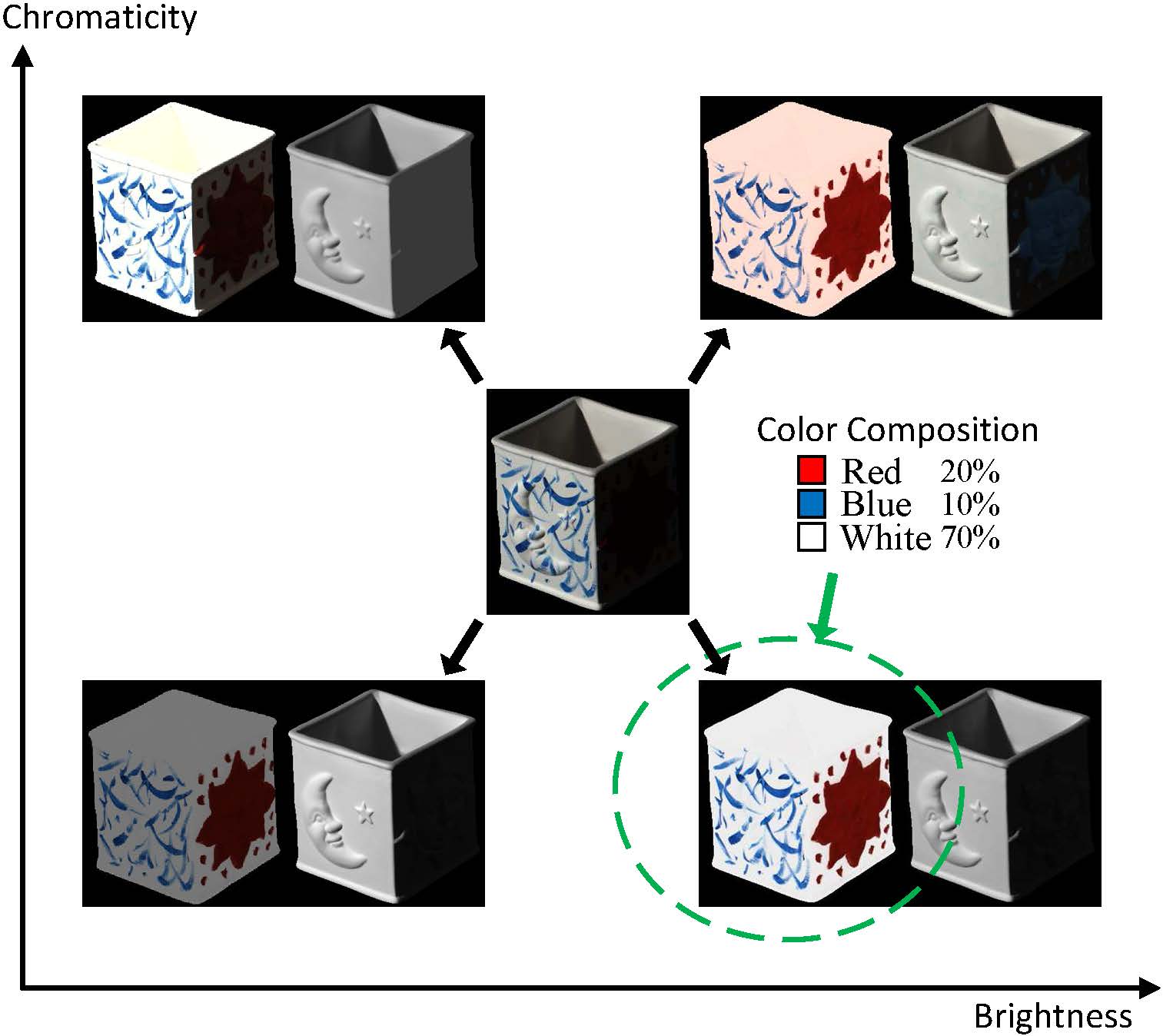}
\end{center}
   \caption{Utility of color composition for intrinsic image decomposition. Different combinations of shading and reflectance can reproduce the input image (in the center of the figure). With the help of color naming, the brightness and the chromaticity of the reflectance can be limited to a certain scope (the green ellipse), so the ambiguity of intrinsic image decomposition can be greatly reduced.}
\label{fig:utility_of_CN}
\end{figure}

In this paper we propose a new type of user interaction that takes color composition of the image as input.
Unlike the traditional guidance on the reflectance and shading of local areas, color composition helps to determine the overall distribution of the reflectance directly. In essence, color composition tells the proportions of colors in the target reflectance. Fig.\ref{fig:utility_of_CN} illustrates the utility of color composition in disambiguating intrinsic image decomposition. These constraints help cleaning the residual of shadows left in the recovered reflectance (which can be seen in the top left part of Fig. \ref{fig:utility_of_CN}), such that the color of shadowed regions will not be mistaken for ``black''. It can also help determining a reasonable illuminant of the scene, so the global scalar of the recovered reflectance and the chromaticity of the shading will be appropriate. After the reflectance is recovered, we can obtain a more detailed description of the colors than the image-level annotation.

We represent the color composition by a distribution over the eleven basic color terms \cite{basiccolor}. In English they are black, blue, brown, gray, green, orange, pink, purple, red, white, and yellow. People are familiar with these concepts, no matter what language they are speaking, so they can describe the color composition quickly and consistently.

We built a generative model that connects the intrinsic images and the color composition through the albedo of materials. The underlying assumption is that the albedo of materials generates both the reflectance in imaging and the color composition in color naming. Based on the global sparsity of reflectance \cite{color_lines,Shen_PAMI13}, we model the albedo of the materials in the scene by a Gaussian Mixture Model, and the intrinsic reflectance can be regarded as an observation of the albedo. There are also strong evidences for the close relation between the color composition and the reflectance.
The physical attribute that primarily determines the perceived color of a surface is exactly its reflectance \cite{Color_vision}. Humans are good at intrinsic image decomposition \cite{Bell_Siggraph14}, although not perfect \cite{Adelson00}. Human vision also shows some degree of color constancy \cite{HumanColorConstancy}. Therefore the annotation of the color composition will not be affected much by the shading or the illumination.

\subsection{Related Work}
The relation between the color composition and the reflectance has already been addressed in the literature.
Our prior work \cite{Color_Naming} demonstrated that color naming on the reflectance gave more accurate color labels than on the raw image.
There are also some early attempts to use color names for intrinsic image decomposition. \cite{Serra_CVPR12} take color-name descriptors as top-down intervention for intrinsic image decomposition. They obtain the pixel-level color names by chip-based method \cite{colorprob}, and unify the color names of pixels that have the same reflectance. These color names are then mapped to RGB values that are supposed to be the reflectance. The problem is that chip-based methods are unaware of the shading and illumination at all, and unifying the color names of pixels with the same reflectance cannot eliminate the effect of shading either. Hence their color composition does not necessarily correspond to the reflectance. In comparison, human annotation of color composition can reduce the influence of shading and illumination considerably.
\cite{IID_Obj_Attr} built a unified framework to jointly estimate the intrinsic images, objects and attributes (including color names) . They verified that utilizing the correlations between intrinsic images and attributes do improve the performance of sub-tasks.

Our model is based on the model of automatic intrinsic image decomposition proposed by \cite{gehler11nips}. Their model is written in the form of Conditional Random Field , taking the smoothness of shading, the sparsity of reflectance, and the Retinex term as additive energy terms. Apart from these common terms, our model differs from the model of \cite{gehler11nips} in the following aspects:
\begin{itemize}
\item Our model utilized the top-down information from color naming to guide the clustering of reflectance, while their clustering are totally data-driven;
\item We addressed the direct and ambient illuminations explicitly, and solved the problem of color constancy;
\item We use Gaussian mixture model instead of k-means to represent the clusters of reflectance, which facilitates the fusion of color naming guidance and data mode in a probabilistic framework.
\end{itemize}

Recently, \cite{chang14svdpgmm} proposed a model based on Dirichlet process Gaussian mixture model. In this model the image was treated as an observation from a generative, stochastic process, and the shading was modeled by a Gaussian process. One superiority of this model is that the optimal number of components for the reflectance image can be decided automatically. However the complexity of this model makes it hard to be integrated with the color naming guidance. In this work we use the Gaussian mixture model directly to represent the surface albedos.

\section{Generative Model of Image and Color Composition}
\label{sec:CN_IID_model}

\begin{figure}[!t]
\begin{center}
\includegraphics[width=0.4\linewidth]{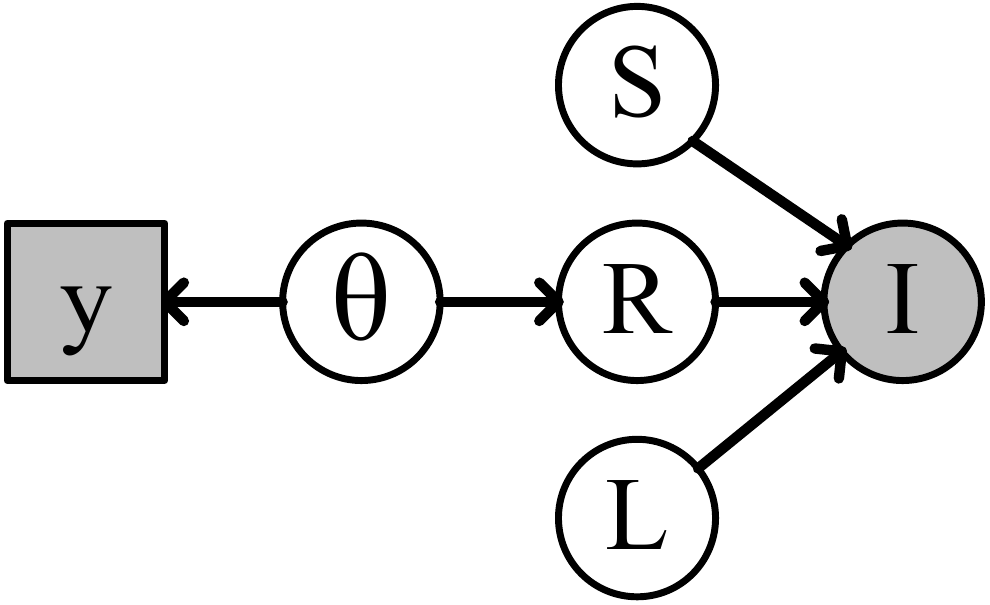}
\end{center}
   \caption{The generative model of image and color composition. The image $\I$ is observable, and its color composition $\y$ is supplied by user annotation. The unobservable variables include the albedo of materials $\bm\theta$, the reflectance $\R$, the shading $\S$, and the global illumination $\L$.}
\label{fig:model}
\end{figure}

We build a generative model of the image $\I$ and its color composition $\y$ as well as the intrinsic images, including the shading $\S$, the reflectance $\R$, and the illumination $\L$. The sketch of our model is depicted in Fig. \ref{fig:model}. The core of our model is the albedo of the materials $\bm\theta$ in the scene, which generates not only the reflectance $\R$ but also the color composition $\y$ of the image in a color naming process.

Given the image and the color composition, the unobserved variables in our model can be determined by solving a MAP problem, where the posterior probability is defined as follows:
\begin{equation}
\label{eqn:CN_IID_crf}
\begin{aligned}
&P(\S,\R,\bm\theta,\L|\I,\y) \propto \\
&\ \ P(\L)P(\S)P(\bm\theta)P(\y|\bm\theta)P(\R|\bm\theta)P(\I|\R,\S,\L).
\end{aligned}
\end{equation}
The probabilities $P(\L)$, $P(\S)$, and $P(\bm\theta)$ are the priors on the illumination, the shading, and the albedo of materials, respectively. The conditional probabilities $P(\y|\bm\theta)$ and $P(\R|\bm\theta)$ are the likelihoods of observing the color composition $\y$ and the reflectance $\R$ on materials $\bm\theta$, respectively. The likelihood of the image $P(\I|\R,\S,\L)$ is often expressed in a deterministic way, \ie $\I=\R \cdot \L \cdot \S$.

Our model has two key features. Firstly, the color composition is involved in our model. The image-level color composition affords informative cues on the colors of individual materials together with their mixture proportions, which will further shape the overall distribution of the reflectance.

Secondly, the global illumination is separated from the reflectance and the shading. Traditionally the illumination is either combined into the shading \cite{BousseauTOG2009,chang14svdpgmm,Tappen_05} or ignored \cite{gehler11nips}. \cite{BarronECCV12} showed that separating the illumination can make the decomposition more complete and meaningful. In our former work \cite{Shading_order} the color constancy problem was avoided by defining the reflectance to be the fully lighted image that all the regions are covered by full direct illumination. In this case the reflectance is actually the illumination-modulated reflectance $\tilde{\R}=\R\cdot \L$. Here we have to separate the illumination from the reflectance, since the perceived color compositions are actually describing the reflectance while the illumination should have been ``whitened'' by the human visual system.

\subsection{Representations of Major Components}
\label{sec:CN_IID_image_formation}


The color composition is defined to be a vector $\y=[\y_1,...,\y_M]^T$, each dimension of which denotes the proportion of one color within the whole image. Specifically, we choose the colors from eleven basic color terms \cite{basiccolor}, so $M=11$. Note that we always have $\|\y\|_1=1$.

Following our former work \cite{Shading_order}, we represent the raw images and the intrinsic images in the $UVB$ color space. This color space is formed by a 2D shadow-free plane $UV$ \cite{BIDR} and a brightness dimension $B$ that depends on the shading.
The image in the $UVB$ space is obtained from rotating the $\log$ RGB space as follows:
\begin{equation}
\small
\label{eqn:CN_IID_UVB}
[I^u_p,I^v_p,I^b_p]=\ln(\I_p) H,
\end{equation}
where $p$ is the index of pixel. The rotation matrix $H=[\u,\v,\n]$ is mainly determined by the brightening direction $\n$, while $\u$ and $\v$ are merely an arbitrary pair of basis vectors of the $UV$ plane perpendicular to $\n$.

Denote the illumination-modulated reflectance in the $UVB$ space by $\tilde{\R}_p=[R^u_p,R^v_p,R^b_p]$. According to the property of shadow-free plane $UV$ \cite{BIDR}, we have the following approximations:
\begin{equation}
\label{eqn:CN_IID_R_uv}
\begin{aligned}
R^u_p &\approx I^u_p = \ln \I_p\cdot \u \\
R^v_p &\approx I^v_p = \ln \I_p\cdot \v.
\end{aligned}
\end{equation}
The only unknown dimension of the illumination-modulated reflectance is the reflectance brightness $R^b_p$, which is the main goal of our inference in Section \ref{sec:CN_IID_inference}. Once we obtain $R^b_p$, we can recover the illumination-modulated reflectance in the RGB space by:
\begin{equation}
\label{eqn:CN_IID_recover_R}
\tilde{\R}_p = exp([R^u_p,R^v_p,R^b_p]H^{-1}),
\end{equation}
where $exp$ denotes element-wise exponential.

According to the definition of the illumination-modulated reflectance, the reflectance can be calculated by $\R_p = \tilde{\R}_p/\L$. In the $UVB$ color space the reflectance will be:
\begin{equation}
\label{eqn:CN_IID_body_refl}
\begin{aligned}
\hat{\R}_p &= [R^u_p,R^v_p,R^b_p]-[L^u,L^v,L^b]\\
           &=[I^u_p-L^u,I^v_p-L^v,R^b_p-L^b],
\end{aligned}
\end{equation}
where $[L^u,L^v,L^b]$ is the illumination in the $UVB$ space. Here we utilized the approximations of $R^u$ and $R^v$ in Equation (\ref{eqn:CN_IID_R_uv}). Note that the illumination in the $UVB$ space acts as a global bias, which does not affect the shape of the distribution of pixels.

\begin{figure}
\begin{center}
\includegraphics[width=1.0\linewidth]{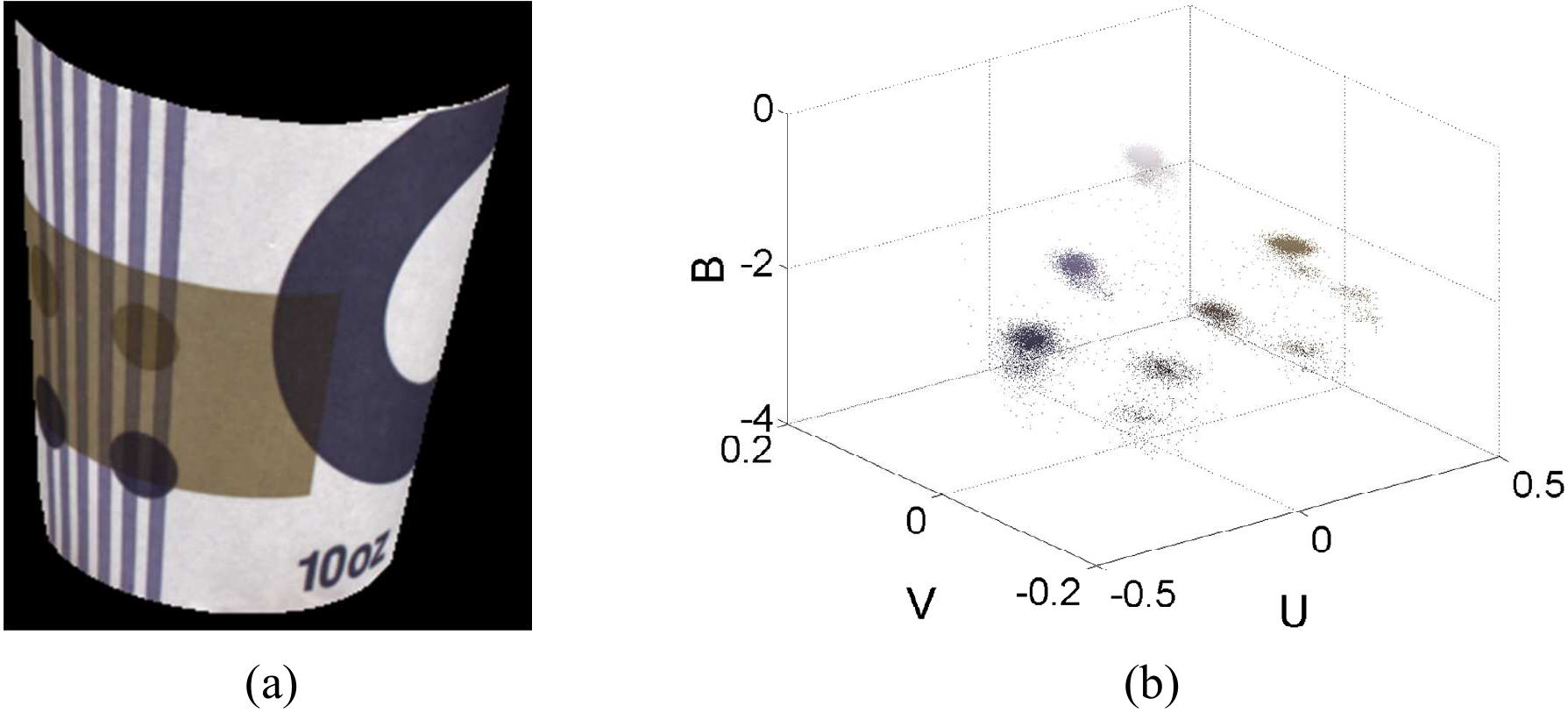}
\end{center}
   \caption{Clusters of reflectance in the $UVB$ space. (a) The reflectance. (b) The clusters. The number of clusters are quite limited, and the shapes of the clusters are Gaussian-like.}
\label{fig:cluster_UVB}
\end{figure}

Most natural images can be represented by a limited number of color lines \cite{color_lines}, each of which corresponding to a material with unique color. After removing the variance of brightness caused by shading, the reflectance of pixels falls naturally into a limited number of compact clusters in the $UVB$ color space as shown in Fig. \ref{fig:cluster_UVB}.
Therefore we build a Gaussian mixture model (GMM) for the albedo of materials $\bm\theta$ in the $UVB$ color space. The GMM of $\bm\theta$ is specified by the means $\bm\mu$, the covariances $\bm\Sigma$, and the mixing coefficients $\bm\pi$. Each Gaussian component is supposed to cover the pixels from a certain material, whose albedo, covariance of color, and population are recorded in the corresponding component of $\bm\mu$, $\bm\Sigma$ and $\bm\pi$, respectively.

\textbf{Discussions.} The $UVB$ color space addresses the bi-illumination scenario \cite{BIDR}, where there is a direct illuminant and an ambient illumination in the scene. The ambient illuminations commonly exist in natural scenes. Modern cameras have dynamic range high enough to capture the reflections of ambient illuminations, even the weak ones such as those appear in the shadows of Fig. \ref{fig:utility_of_CN}.

The major advantage of adopting the $UVB$ color space is that the chromaticity and the intensity can be separated in the bi-illumination scenario, while separating chromaticity from intensity is critical for reducing the number of unknown variables of intrinsic images \cite{Dominant_color,gehler11nips,Garces2012}. This property derives the approximations in Equation \ref{eqn:CN_IID_R_uv}. Some traditional color spaces, such as normalized RGB \cite{Dominant_color,gehler11nips} and Lab \cite{Garces2012}, are also capable of chromaticity-intensity separation. However, their accuracies are unguaranteed in the bi-illumination scenario, since shading often changes the dimensions of chromaticity in these color spaces as well \cite{Dominant_color}. The $UVB$ color space increases the accuracy by determining the brightening direction $\n$ from data, instead of predefining it to be neutral \cite{intrinsic_dataset}. The brightening direction $\n$ captures the principle direction of color changes caused by shading variations, so the null space of $\n$ will be mostly shadow-free.


\subsection{Objective Function}

The MAP problem in Equation (\ref{eqn:CN_IID_crf}) is transformed into an energy minimization problem as usual. The posterior probability is realized by the following energy function:
\begin{equation}
\label{eqn:CN_IID_energy}
\begin{aligned}
E(R^b,\bm\theta,\L;\I,\y) = &w_s E_s(R^b) + w_r E_{r}(R^b) + w_g E_g(\bm\theta)\\
                            &+ E_{d}(R^b,\bm\theta,\L;\I) + w_c E_c(\bm\theta;\y)\\
\mbox{s.t. }\hspace{3mm}\L\in\mathcal{L},\hspace{3mm} &\|\bm\pi\|_1 = 1, \hspace{3mm}\bm\pi_k\geq 0, \forall k,
\end{aligned}
\end{equation}
where $w_s$, $w_r$, $w_g$ and $w_c$ are the weights of the energy terms. The illumination $\L$ is assumed to be uniformly distributed within the feasible domain $\mathcal{L}$, so $P(\L)$ in Equation (\ref{eqn:CN_IID_crf}) can be omitted. The optimal illumination is selected to be the one that makes the color composition of reflectance consistent with the annotation. The energy terms are described in the following paragraphs.

The smoothness of shading $E_s$ and the Retinex term $E_r$ are similar to those used by
\cite{gehler11nips} but formulated in the $UVB$ space:
\begin{equation}
\label{eqn:CN_IID_E_s_UVB}
E_s(R^b) = \sum_{p\sim q}(R^b_p-R^b_q-(I^b_p-I^b_q))^2,
\end{equation}
and
\begin{equation}
\label{eqn:CN_IID_E_r_UVB}
E_r(R^b) = \sum_{p\sim q}(R^b_p-R^b_q-g_{p,q}(\I)\cdot(I^b_p-I^b_q))^2,
\end{equation}
where $p\sim q$ denotes the neighborhood relation in a 4-connected pixel graph. $g_{p,q}(\I)\in \{1,0\}$ indicates whether the edge between pixels $p$ and $q$ is a reflectance edge or not. Following Color Retinex \cite{intrinsic_dataset} we calculate the gradients in chromaticity ($UV$ dimensions) and brightness. If the magnitude of the gradient of chromaticity is greater than a threshold $T^{c}$ or the gradient of brightness is greater than a threshold $T^{bl}$ but lower than a threshold $T^{bu},$\footnote{The gradient of brightness at reflectance edges are often not as large as deep shadow edges \cite{Color_vision}.} we classify the edge to be a reflectance edge. The thresholds are chosen by cross-validation. The shading brightness $S^b$ is replaced by $I^b-R^b-L^b$.
If there is no reflectance edge between pixel $p$ and its neighbor $q$, the shading order $S^b_p-S^b_q$ should be equal to their brightness order $I^b_p-I^b_q$. Otherwise the shading order is expected to be around 0 due to the smoothness of shading. Note that $E_s$ is a special case of $E_r$ when $g_{p,q}(\I)=1$. Nevertheless $E_s$ is still useful to suppress the sharp shading changes where the reflectance edges are not detected out.

The prior of the global sparsity of reflectance is realized by minimizing the variances of the clusters:
\begin{equation}
\label{eqn:CN_IID_E_g}
E_g(\bm\theta) = \sum_k \ln |\bm\Sigma_k|,
\end{equation}
where $|\cdot|$ is the determinant of matrix. Note that, the covariance matrices here are specified to be diagonal, and they are always positive definite.

The reflectance term is represented by the negative log likelihood of reflectance under the GMM model:
\begin{equation}
\label{eqn:CN_IID_log_likelihood}
\begin{aligned}
E_{d}(R^b,\bm\theta,\L;\I) &= - \ln P(\hat{\R}|\bm\theta)\\
                           &= -\sum_p \ln\{\sum_k \bm\pi_k \mathcal{N}(\hat{\R}_p|\bm\mu_k, \bm\Sigma_k)\},
\end{aligned}
\end{equation}
where the reflectance $\hat{\R}$ in the $UVB$ space is calculated by Equation (\ref{eqn:CN_IID_body_refl}). To decouple the variables of different components, a binary matrix $\z=\{\z_{p,k}\}_{(p,k)=(1,1)}^{N,K}$ is introduced to denote the assignments of the $N$ pixels to the $K$ clusters \cite{PRML}. Define the probability of assignment as $\gamma(\z_{p,k})=P(\z_{p,k}=1|\hat{\R}_p)$. The reflectance term can be rewritten to be:
\begin{equation}
\label{eqn:CN_IID_log_likelihood_z}
E_{d}(R^b,\bm\theta,\L;\I)=-\sum_{p,k} \gamma(\z_{p,k}) \{\ln \bm\pi_k + \ln \mathcal{N}(\hat{\R}_p|\bm\mu_k, \bm\Sigma_k)\}.
\end{equation}

The color naming term is modeled by:
\begin{equation}
\label{eqn:CN_IID_color_constancy}
E_c(\bm\theta;\y) = \|\y - \tilde{\y}(\bm\mu)\bm\pi \|_2^2,
\end{equation}
where $\tilde{\y}(\bm\mu)=[\tilde{\y}(\bm\mu_1),...,\tilde{\y}(\bm\mu_K)]$ is the component-level color composition. The color composition of the $k$-th component of the GMM is computed by $\tilde{\y}(\bm\mu_k)=T(exp(\bm\mu_k\cdot H^{-1}))$, where $T$ is the color naming projection from the RGB space to the color name space. In our experiments we adopt the method of \cite{colorprob}, which is a well-known chip-based color naming method. Here we apply color naming on the cluster centers instead of individual pixels, which saves a lot of computation cost. Color naming on pixel clusters seems more likely to be the way that people determine the color composition of images. The color assimilation effect will promote the perceived color of clustered pixels to be the same \cite{ColorAssi}. Therefore the variances of colors within the clusters are unlikely to change the color naming results. Here the image-level color composition of the reflectance is the weighted sum of the component-level color compositions, while the weights are the mixing coefficients of the GMM.

\textbf{Discussions.} The color naming term is the key to realize color naming guided intrinsic image decomposition. Minimizing the energy $E_c$ with respect to $\bm\pi$ will adapt the mixture coefficients of the GMM until the accumulation of component-level color compositions approximates the image-level color composition. After that, the assignments of the pixels to the components will be updated in a way that the larger the mixture coefficient is, the larger the population of the corresponding Gaussian component will be. Especially, the components formed by shadowed pixels will be gradually merged into the components of normally lighted pixels, and their brightness will be raised accordingly during the minimization of the reflectance term $E_d$ in Equation (\ref{eqn:CN_IID_log_likelihood_z}).


\section{Inference}
\label{sec:CN_IID_inference}

\begin{figure*}[t]
\begin{center}
\includegraphics[width=1.0\linewidth]{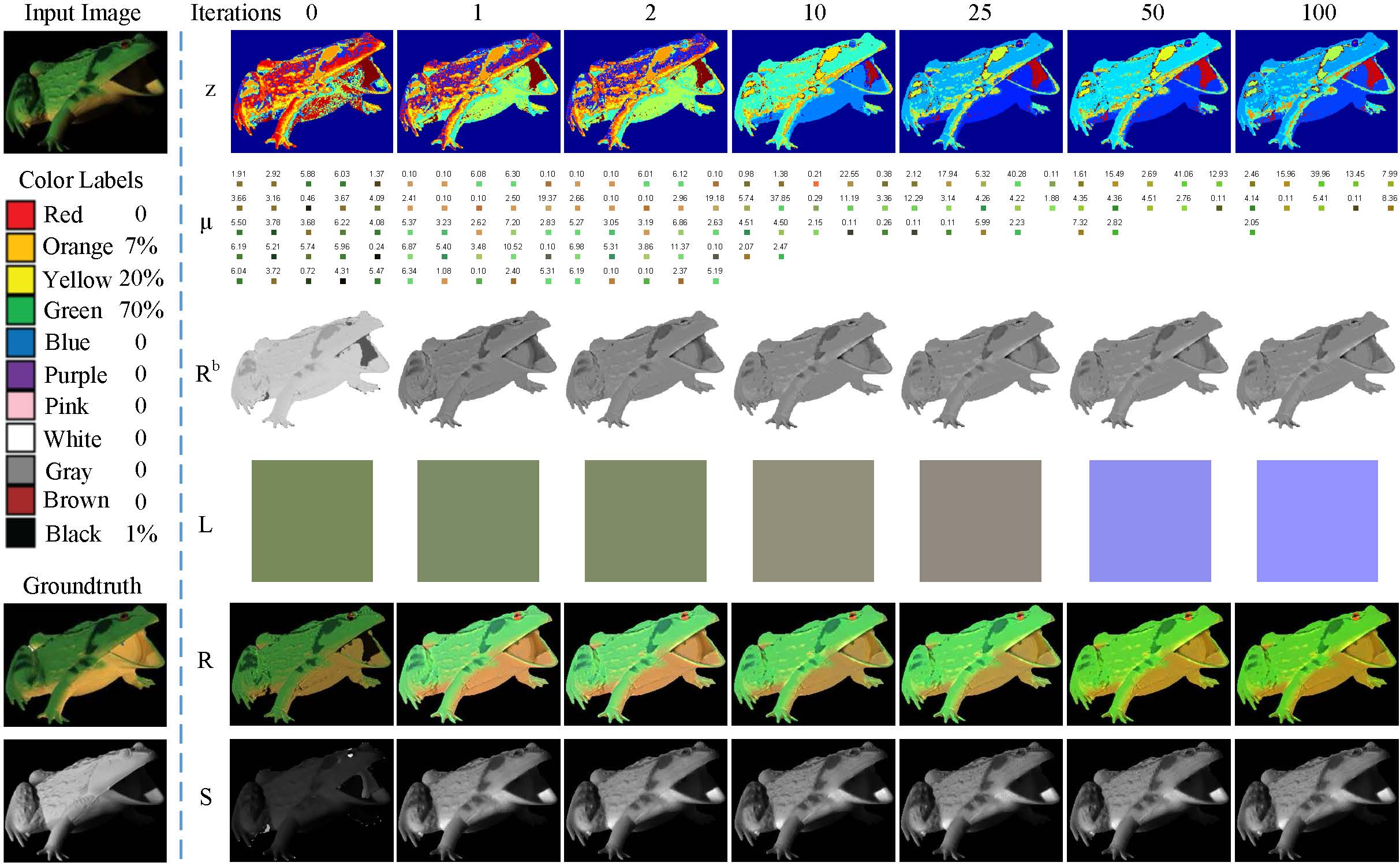}
\end{center}
   \caption{Temporal results of the iterative optimization.}
\label{fig6_3:iteration}
\end{figure*}

We minimize the energy function in Equation (\ref{eqn:CN_IID_energy}) in an iterative way. The whole process is summarized in Algorithm \ref{alg:CN_IID_optimization}. The key operation is to search the optimal mixing coefficients $\bm\pi$, which not only explain the distribution of the components in the GMM but also reconstruct the image-level color composition from the component-level color compositions. The temporal results of an example image in the iterative procedure are shown in Fig. \ref{fig6_3:iteration}.

\begin{algorithm}[t!]
\caption{Color Naming guided Intrinsic Image Decomposition}
\label{alg:CN_IID_optimization}
\begin{algorithmic}[l]
\REQUIRE
The raw image $\I$, the color composition $\y$, and the threshold of energy drop $\delta$.
\ENSURE
$R^b$ and $\L$.
\STATE Initialize $R^b_0$ as described in the text. The other two dimensions of $\tilde{\R}_0$ are inherited from the image according to Equation (\ref{eqn:CN_IID_R_uv}).
\STATE Fit GMM $\tilde{\bm\theta}_0$ over $\tilde{\R}_0$ by the standard EM algorithm \cite{PRML}.
\STATE $\bm\pi_0\leftarrow \tilde{\bm\pi}_0$, $\bm\Sigma_0\leftarrow \tilde{\bm\Sigma}_0$,
\STATE Search the optimal $\L_0$ that makes $\bm\mu_0=\tilde{\bm\mu}_0-\L_0$ minimize the color naming term in Equation (\ref{eqn:CN_IID_color_constancy}).
\STATE $t\leftarrow 0$.
\REPEAT
        \STATE $R^b_{t+1}\leftarrow$ Optimize $R^b$ using Equation (\ref{eqn:CN_IID_opt_r});
        \STATE $\L_{t+1}\leftarrow$ Select $\L$ using Equation (\ref{eqn:CN_IID_infer_L});
        \STATE $\bm\theta_{t+1}\leftarrow$ Update $\bm\theta$ using Equations (\ref{eqn:CN_IID_opt_gamma}), (\ref{eqn:CN_IID_update_GMM}), and (\ref{eqn:CN_IID_opt_Pi});
        \STATE $t\leftarrow t+1$;
\UNTIL{$E(R^b_{t},\bm\theta_{t},\L_{t};\I,\y)-E(R^b_{t-1},\bm\theta_{t-1},\L_{t-1};\I,\y)<\delta$}
\RETURN $R^b_{t}$ and $\L_{t}$.
\end{algorithmic}
\end{algorithm}

\noindent\textbf{Optimize $R^b$} by:
\begin{equation}
\label{eqn:CN_IID_opt_r}
R^b_{t+1} = \arg\min_{R^b} w_s E_s(R^b) + w_r E_{r}(R^b) + E_{d}(R^b,\bm\theta_{t},\L_{t};\I).
\end{equation}
All the energy terms are differentiable, so the optimization can be solved efficiently by gradient descend \cite{GradientDescent}.

\noindent\textbf{Select illumination $\L$} by:
\begin{equation}
\label{eqn:CN_IID_infer_L}
\L_{t+1}=\arg \min_{\L\in \mathcal{L}} E_d(R^b_{t+1},\bm\theta_{t},\L;\I).
\end{equation}
The feasible domain of illumination $\mathcal{L}$ covers the chromaticity of all the illumination of the dataset, while the intensity is empirically set to be within $[0.5,2]$. The feasible domain is then sampled uniformly, and the optimal illumination is selected from those samples.

\noindent\textbf{Update GMM $\bm\theta$}. We minimize the energy $E(R^b,\bm\theta,\L;\I,\y)$ in Equation (\ref{eqn:CN_IID_energy}) with respect to the parameters $\bm\theta$ of GMM by the EM algorithm. In the E step, we evaluate the probability of the assignments of the $N$ pixels to the $K$ clusters by:
\begin{equation}
\label{eqn:CN_IID_opt_gamma}
\gamma(\z_{p,k})=\frac{\bm\pi_k \mathcal{N}(\hat{\R}_p|\bm\mu_k, \bm\Sigma_k)}{\sum_j \bm\pi_j \mathcal{N}(\hat{\R}_p|\bm\mu_j, \bm\Sigma_j)}.
\end{equation}
In the M step, we first update the mean $\bm\mu$ and the variance $\bm\Sigma$ as follows:
\begin{equation}
\label{eqn:CN_IID_update_GMM}
\begin{aligned}
\bm\mu_k &= \frac{1}{\N_k}\sum_p \gamma(\z_{p,k}) \hat{\R}_p\\
\bm\Sigma_k &= \frac{1}{\N_k+2w_g}\sum_p \gamma(\z_{p,k})(\hat{\R}_p-\bm\mu_k)(\hat{\R}_p-\bm\mu_k)^T,
\end{aligned}
\end{equation}
where $\N_k = \sum_p \gamma(\z_{p,k})$ is the population of the $k$-th component. It should be pointed out that we do NOT minimize $E_c$ with respect to $\bm\mu$, since the color naming function such as that defined by \cite{colorprob} is too complex to be optimized analytically. Through experiments we found that this simplification did not change the results much. The change of $\bm\mu$ in a single iteration is usually too small to affect color naming. Note that the formulation of $\bm\Sigma_k$ in Equation (\ref{eqn:CN_IID_update_GMM}) is slightly different from those in the general EM algorithm for fitting GMM \cite{PRML}, with an extra term $w_g$ in the denominator that reduces the variance. The reason is that the sparsity of reflectance (defined as $E_g$ in Equation (\ref{eqn:CN_IID_E_g})) is incorporated into the energy function.
\begin{figure*}[t]
  \centering
  \includegraphics[width=.9\linewidth]{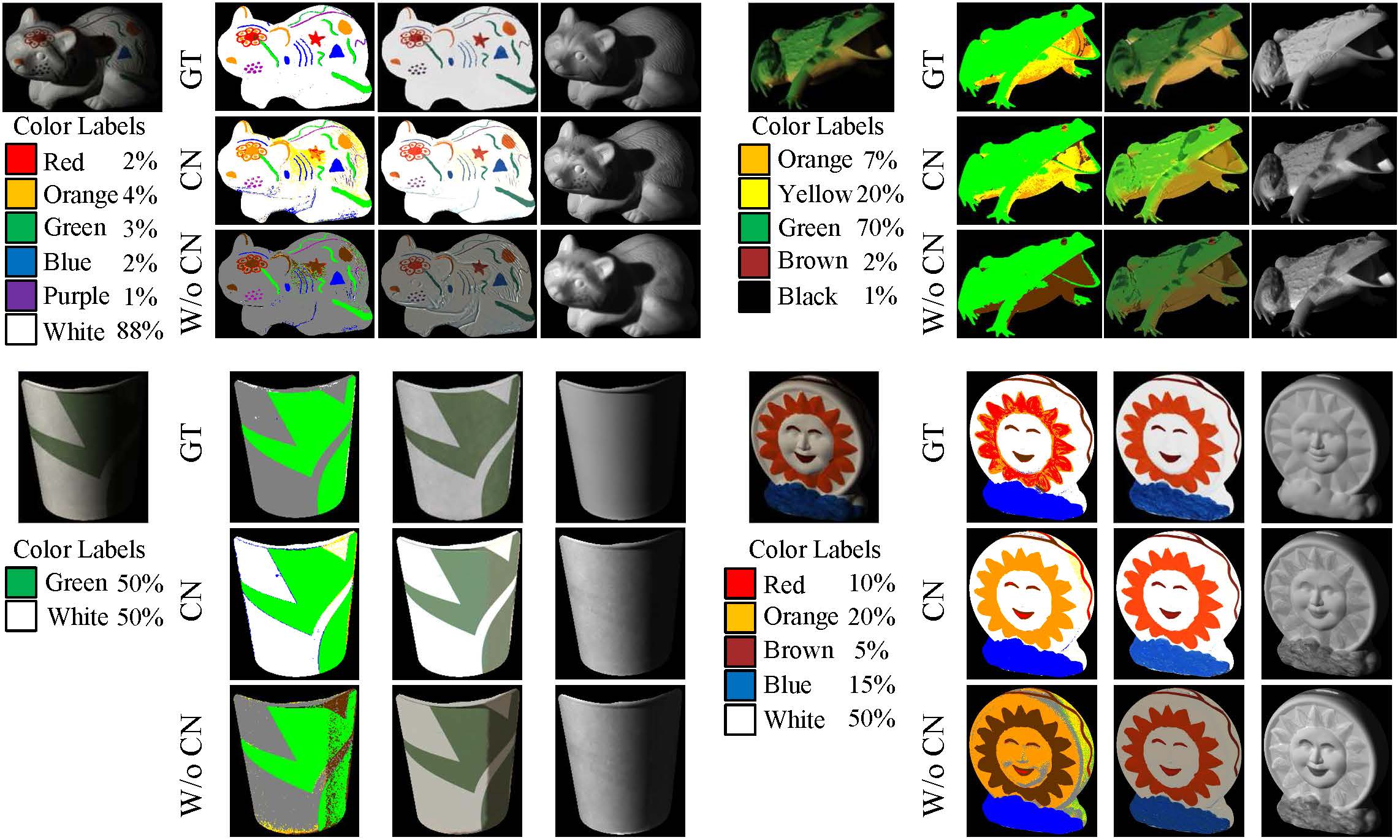}\\
  \caption{Comparison of results with or without color naming as guidance. \textbf{GT}: Groundtruth intrinsic images and the color naming results over the reflectance. \textbf{CN}: The results of our method taking color naming as guidance. \textbf{W/o CN}: The results of our method without color naming.}\label{fig:CN_IID_CN_vs_noCN}
\end{figure*}

The optimal mixture coefficient $\bm\pi$ is obtained by:
\begin{equation}
\label{eqn:CN_IID_opt_Pi}
\begin{aligned}
&\arg \min_{\bm\pi} -\N^T \ln \bm\pi + w_c\|\y - \tilde{\y}(\bm\mu)\bm\pi \|_2^2 \\
&\mbox{s.t. } \|\bm\pi\|_1=1, \ \bm\pi_k \geq 0, \forall k.
\end{aligned}
\end{equation}
We solve this problem by the Alternating Direction Method of Multipliers (ADMM) \cite{ADMM}. We reformulate the problem in Equation (\ref{eqn:CN_IID_opt_Pi}) as follows:
\begin{equation}
\label{eqn:CN_IID_augmented_L}
\begin{aligned}
&\arg\min_{\bm\pi,\bm\phi,\bm\psi} f(\bm\pi)+g(\bm\phi)+h(\bm\psi)\\
&\mbox{s.t. } \|\bm\pi\|_1=1\\
&\ \ \ \ \ \   \bm\pi=\bm\phi=\bm\psi,
\end{aligned}
\end{equation}
where
\begin{equation}
\label{eqn:CN_IID_g}
\begin{aligned}
f(\bm\pi)&=w_c\|\y - \tilde{\y}(\bm\mu)\bm\pi \|_2^2\\
g(\bm\phi)&=-\N^T \ln \bm\phi\\
h(\bm\psi)&=\left\{
\begin{aligned}
&0  &\mbox{if } \bm\psi_k \geq 0, \forall k \\
&\infty &\mbox{otherwise.}\\
\end{aligned}\right.
\end{aligned}
\end{equation}

Through introducing Lagrange multipliers $\lambda$, $\Gamma$ and $\Upsilon$, we can obtain the following augmented Lagrangian \cite{ADMM}:
\begin{equation}
\begin{aligned}
\label{eqn:L_W}
\mathcal{L}(\bm\pi,\bm\phi,\bm\psi,\lambda,\Gamma,\Upsilon)=& f(\bm\pi)+g(\bm\phi)+h(\bm\psi)+ \lambda (\|\bm\pi\|_1-1) \\
&+ \Gamma^T (\bm\pi-\bm\phi)+\frac{\rho}{2}\|\bm\pi-\bm\phi\|_2^2\\
&+ \Upsilon^T (\bm\pi-\bm\psi)+\frac{\rho}{2}\|\bm\pi-\bm\psi\|_2^2,
\end{aligned}
\end{equation}
where $\rho$ is a scaling parameter, which is empirically set to be 20. We initialize $\bm\pi$, $\bm\phi$, $\bm\psi$, $\Gamma$, and $\Upsilon$ with $\bm\pi_t$, while $\lambda=1$.
Then we update them iteratively as follows:
\begin{equation}
\label{eqn:update_vars}
\begin{aligned}
\bm\pi_{i+1}=&\bm\pi_{i}-\eta \left(2w_c\tilde{\y}(\bm\mu)\tilde{\y}(\bm\mu)^T\bm\pi_{i}-2w_c\tilde{\y}(\bm\mu)\y\right.\\
&\left.+\Gamma_{i}+\rho(\bm\pi_{i}-\bm\phi_{i})
 +\Upsilon_{i}+\rho(\bm\pi_{i}-\bm\psi_{i})+\lambda_{i}\right)\\
\bm\phi_{i+1}=&\bm\phi_{i}-\eta \left(- (diag(\bm\pi_{i+1}))^{-1}\N-\Gamma_{i}-\rho (\bm\pi_{i+1}-\bm\phi_{i})\right)\\
\bm\psi_{i+1}=&(\pi_{i+1}+\frac{1}{\rho}\Upsilon_{i})_+\\
\lambda_{i+1}=& \lambda_{i} + \eta (\|\bm\pi_{i+1}\|_1-1)\\
\Gamma_{i+1}=& \Gamma_{i} + \eta (\bm\pi_{i+1} - \bm\phi_{i+1})\\
\Upsilon_{i+1}=& \Upsilon_{i} + \eta (\bm\pi_{i+1} - \bm\psi_{i+1}),\\
\end{aligned}
\end{equation}
where $(\cdot)_+$ truncates all the elements of a vector to be non-negative. $\eta$ is the step size, which is set to be 0.001 in our experiments. We terminate the iteration when the reduction of the objective function in Equation (\ref{eqn:CN_IID_opt_Pi}) is less than a threshold $T_d$, and we set $T_d$ to be $10^{-6}$ in implementation. The $\bm\psi$ of the last round is taken to be the output $\bm\pi_{t+1}$.

\noindent\textbf{Initialize $R^b$}. We initialize the brightness of reflectance by $R^b=I^b-\hat{S}^b$, where $\hat{S}^b$ is the shifted shading brightness got from compensating the categorical bias of the shading brightness, as described by \cite{Shading_order}. Especially, the number of categories is set to be the number of positive elements in the color composition $\y$. Since the brightness is ignored in the clustering process of the initialization stage, pixels with the same chromaticity but different intensities will be put into the same category. These pixels will be separated into different GMM components in the iterative updates when the other cues such as the smoothness of shading are incorporated.


\section{Experiments}

We evaluate our method on the MIT intrinsic image dataset\footnote {\url{http://dspace.mit.edu/handle/1721.1/59363}} \cite{intrinsic_dataset}, which is a widely used benchmark for intrinsic image decomposition. This dataset contains challenging images with deep shadows. The results are measured by different types of metrics including MSE, LMSE \cite{intrinsic_dataset}, aLMSE and correlation \cite{correlation}. We also test on the images supplied by \cite{BousseauTOG2009}, which are commonly used by user-assisted methods. The color compositions of all the test images are annotated by the authors of this paper, each of which takes only a couple of seconds to annotate.

In our experiments the number of components $K$ in the initial GMM is set to be 25. The weights of energy terms $w_s$, $w_r$, $w_g$, and $w_c$ in Equation (\ref{eqn:CN_IID_energy}) are set to be $10$, $100$, $0.5N/K$, and $5N$, respectively. Here $N$ is the number of pixels. The threshold of energy drop $\delta$ in Algorithm \ref{alg:CN_IID_optimization} is set to be $0.01$.

To process the images in the Intrinsic image dataset, our algorithm takes about eight minutes for each image on average.

\subsection{The Effects of Color Naming Guidance}
\begin{table}
\caption{Results on the MIT Intrinsic Images dataset. Higher correlation, lower MSE, LMSE and aLMSE are better.}
\label{tab:CN_IID_results}
\centering
\begin{tabular}{lcccc} \hline
                                                & Correlation      & MSE        & LMSE              & aLMSE \\ \hline
Gehler \etal       & 0.7748            & 0.0985   &  0.0244      & 0.2544       \\
Chang \etal            & - & - &0.0229 & -\\
Serra \etal      & 0.7862            & 0.0834    &  0.0340      & 0.2958      \\ \hline
IID w/o CN & 0.8378          & 0.0719          & 0.0209          & 0.2221\\
CN-IID     & 0.8489          & 0.0687          & 0.0204          & 0.2152 \\
GtCN-IID   & \textbf{0.8494} & \textbf{0.0676} & \textbf{0.0203} & \textbf{0.2129} \\
\hline
\end{tabular}
\end{table}

Fig. \ref{fig:CN_IID_CN_vs_noCN} illustrates how the color naming guidance changes the results of intrinsic image decomposition. For comparison we execute a version of our method that the color naming term $E_c$ is excluded from the energy in Equation (\ref{eqn:CN_IID_energy}). The most significant improvement brought by color naming guidance is that the overall color composition of the recovered reflectance gets much closer to our color perception. For example, the raccoon in the top left part of Fig. \ref{fig:CN_IID_CN_vs_noCN} has a white body, and color naming guided intrinsic image decomposition (CN-IID) outputs the reflectance in the right color. Without color naming guidance, the recovered reflectance is darker than the groundtruth and the body of the raccoon appears to be gray. The performance on local areas has also been improved notably. For instance, the cup in the bottom left part of Fig. \ref{fig:CN_IID_CN_vs_noCN} has very different shading between the left side and the right side of the image. The color composition plays two important roles in dealing with the deep shadows: (1) cleaning the shadow residual in the reflectance by eliminating the color of brown that does not occur in the color labels; and (2) choosing a proper number of categories for initializing the reflectance brightness $R^b$ (Section \ref{sec:CN_IID_inference}). As a by-product CN-IID can produce pixel-level color naming results after recovering the reflectance. This is a promising way of illumination-robust color naming.

\begin{figure}[t]
  \centering
  \includegraphics[width=1.0\linewidth]{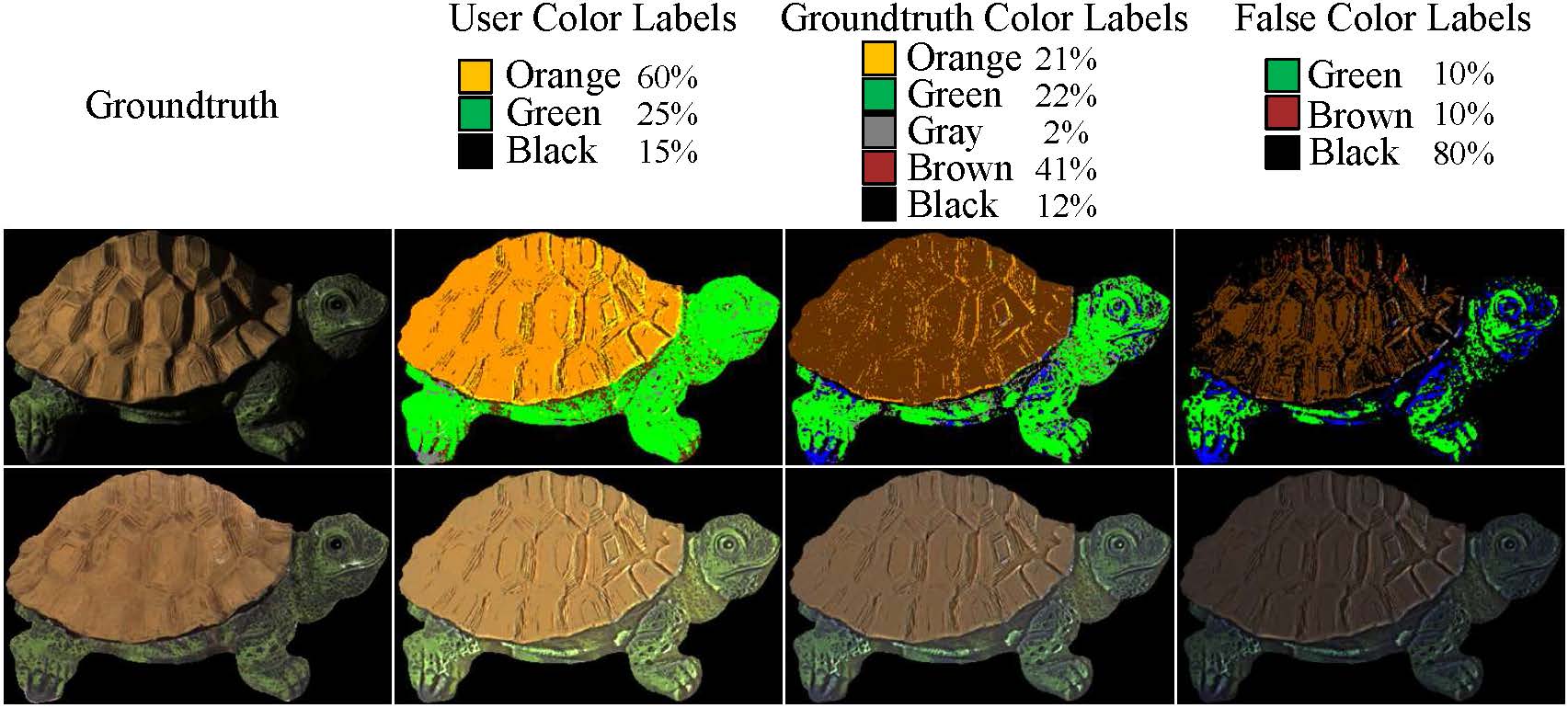}\\
  \caption{Comparison of results under different color naming guidance. Groundtruth color labels are obtained from applying the chip-based color naming method \cite{colorprob} on the groundtruth reflectance.}\label{fig:CN_IID_Different_CN}
\end{figure}

Fig. \ref{fig:CN_IID_Different_CN} demonstrates the reflectance recovered under different color naming guidance. The user labels the shell of the turtle to be orange (the second column of Fig. \ref{fig:CN_IID_Different_CN}), which drives the reflectance to be slightly brighter than the groundtruth. When taking the color composition of the groundtruth reflectance as input (the third column of Fig. \ref{fig:CN_IID_Different_CN}), the shell of the turtle appears to be brown. Both of these outputs are reasonable, and the shadows in local regions are mostly removed from the reflectance. When the input color composition goes wrong, the reflectance will be ruined. For example, in the last column of Fig. \ref{fig:CN_IID_Different_CN} the color of "black" occupies too much proportion of the color composition, so the shadows are left in the reflectance to produce black colors.

The quantitative results under different color naming guidance are given in Table \ref{tab:CN_IID_results}. Generally, the differences are small, since all the metrics are invariant to the global scale of the intensity and the chromaticity of the intrinsic images, which are the major concerns of adopting color naming guidance. Nevertheless, the quantitative results reflect the ability of color naming guidance to improve the performance in local areas, such as removing the residual of shadows in reflectance. Taking the groundtruth color labels as input does perform better than the version that does not take any color naming guidance, referred to as IID w/o CN. The performance when user annotations are taken as input is comparable to the version of GtCN-IID where the groundtruth color labels are given. It suggests that our method is robust to the quality of color naming.

\subsection{Comparison with Automatic Methods}
\begin{figure*}[t]
  \centering
  \includegraphics[width=1.0\linewidth]{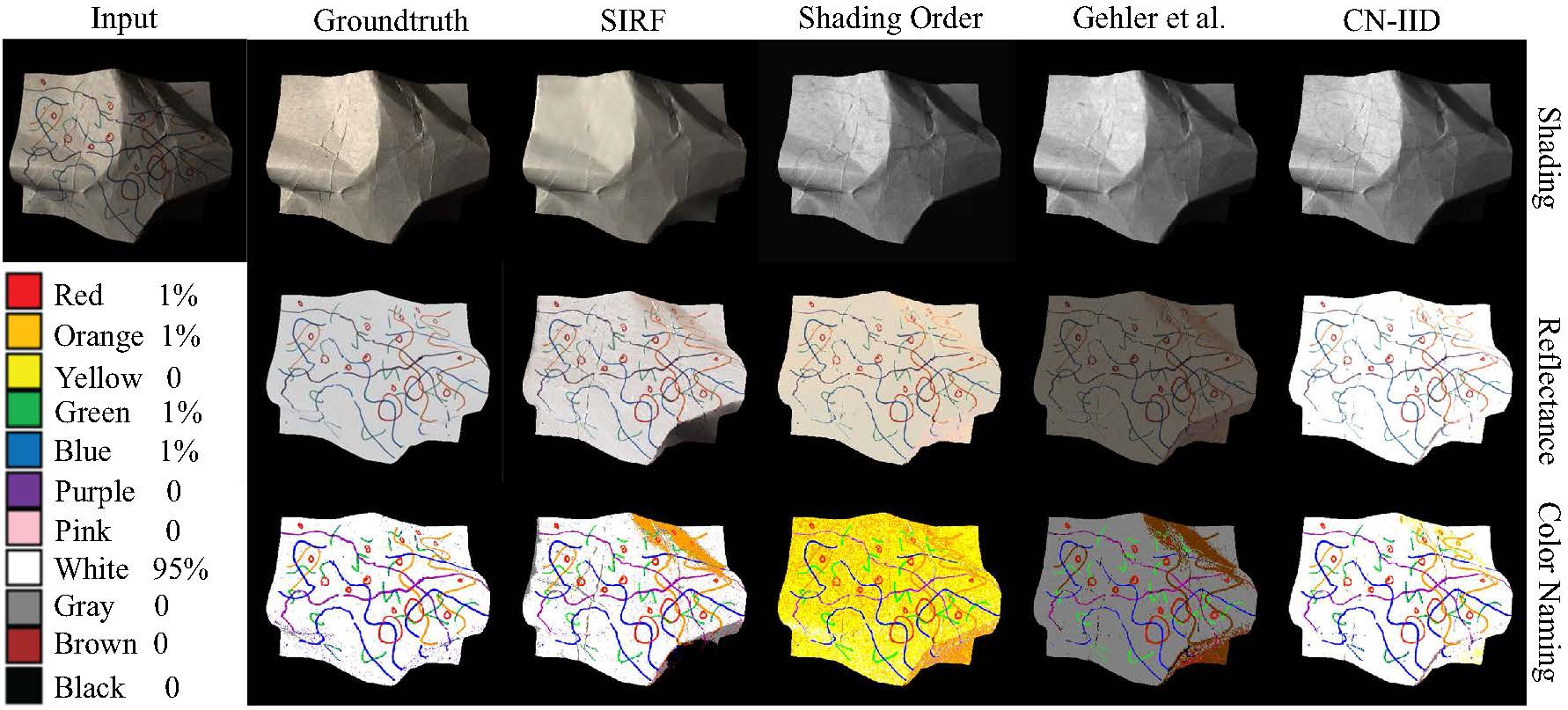}\\
  \caption{Comparison to automatic intrinsic image decomposition. The first column shows the input image (top) and the annotation of color composition (bottom) that is only used by the method of CN-IID. For the other columns, the rows from top to bottom are the output shading, reflectance, and the color composition of reflectance, respectively. The output color labels is produced by mapping the RGB values of reflectance to basic color terms using the method of parametric fuzzy sets \cite{colorprob}. Clearly, the reflectance obtained by CN-IID is visually the most close one to the groundtruth.}\label{fig:CN_IID_motivation}
\end{figure*}

We compare our method to the baseline method of \cite{gehler11nips} and the recent generative model of \cite{chang14svdpgmm}, which also built a GMM model of the reflectance. We also compare to the method of \cite{Serra_CVPR12} that utilizes automatic color naming for intrinsic image decomposition. The results are listed in Table \ref{tab:CN_IID_results}. We can see that CN-IID achieved the best results in all the metrics. It suggested that the top-down information of color composition does facilitate intrinsic image decomposition.

\begin{figure}[!t]
  \centering
  \includegraphics[width=1.0\linewidth]{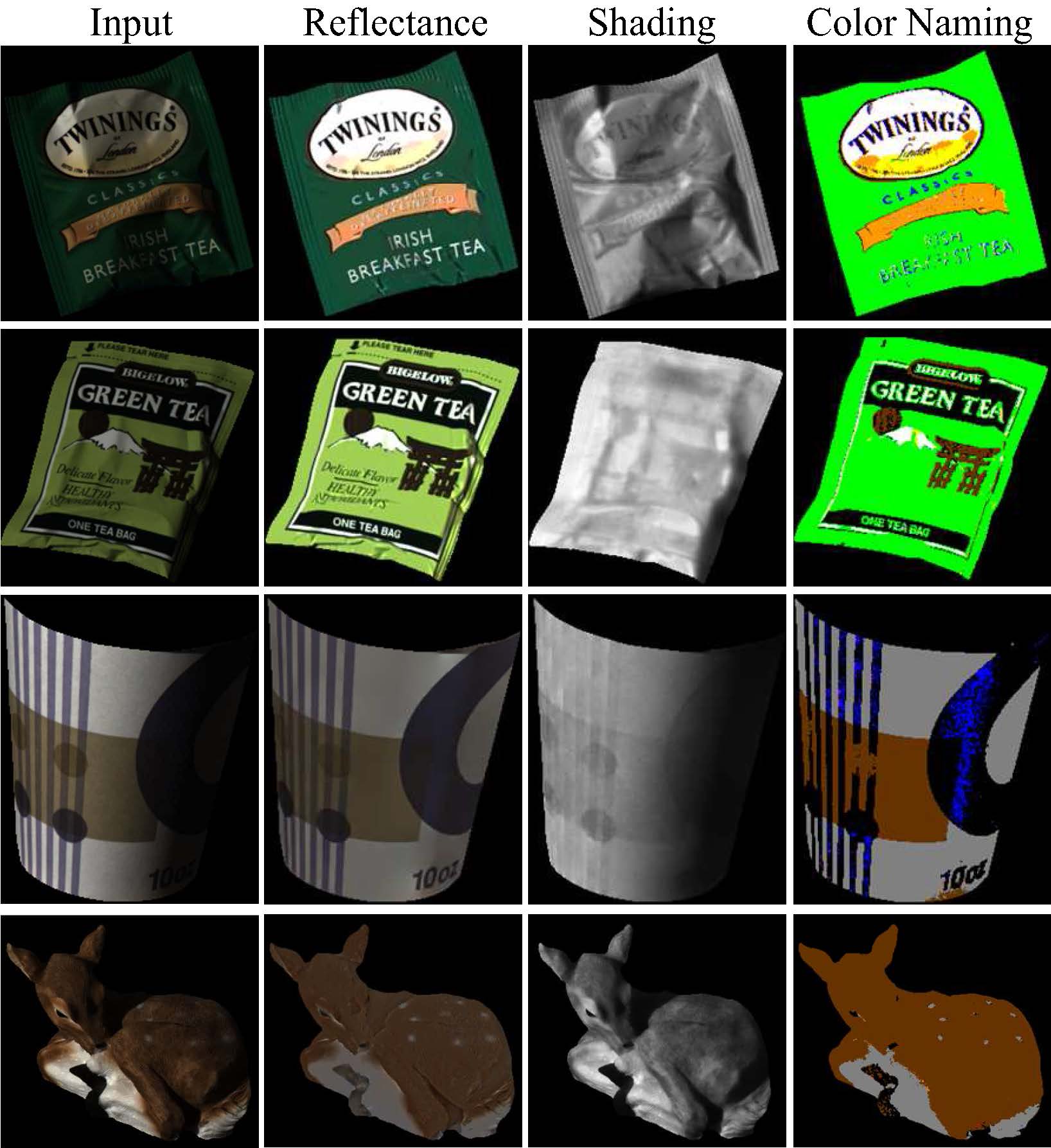}\\
  \caption{Typical results of the MIT intrinsic image dataset.}\label{fig:CN_IID_more_results}
\end{figure}

A qualitative comparison between several representative methods is given in Fig. \ref{fig:CN_IID_motivation}. We can see that the color of the reflectance recovered by CN-IID is most similar to the ground truth. It is worth noting that CN-IID adopted only a rough annotation of the color composition, which is adequate to improve the result considerably. The methods of automatic intrinsic images decomposition suffered various ambiguities. The SIRFS model \cite{BarronECCV12} mistook some shadowed areas as dark surfaces. Our former work based on shading orders \cite{Shading_order} did not remove the chromaticity of the illuminant out of the reflectance, so the white paper turned to be yellow. The method of \cite{gehler11nips}, which is baseline of CN-IID, did not solve the scaling problem, so the estimated reflectance was much darker than it really was. More results obtained by CN-IID on the MIT intrinsic image dataset are given in Fig. \ref{fig:CN_IID_more_results}.

\subsection{Comparison with User-assisted Methods}

\begin{figure*}[t]
  \centering
  \includegraphics[width=1.0\linewidth]{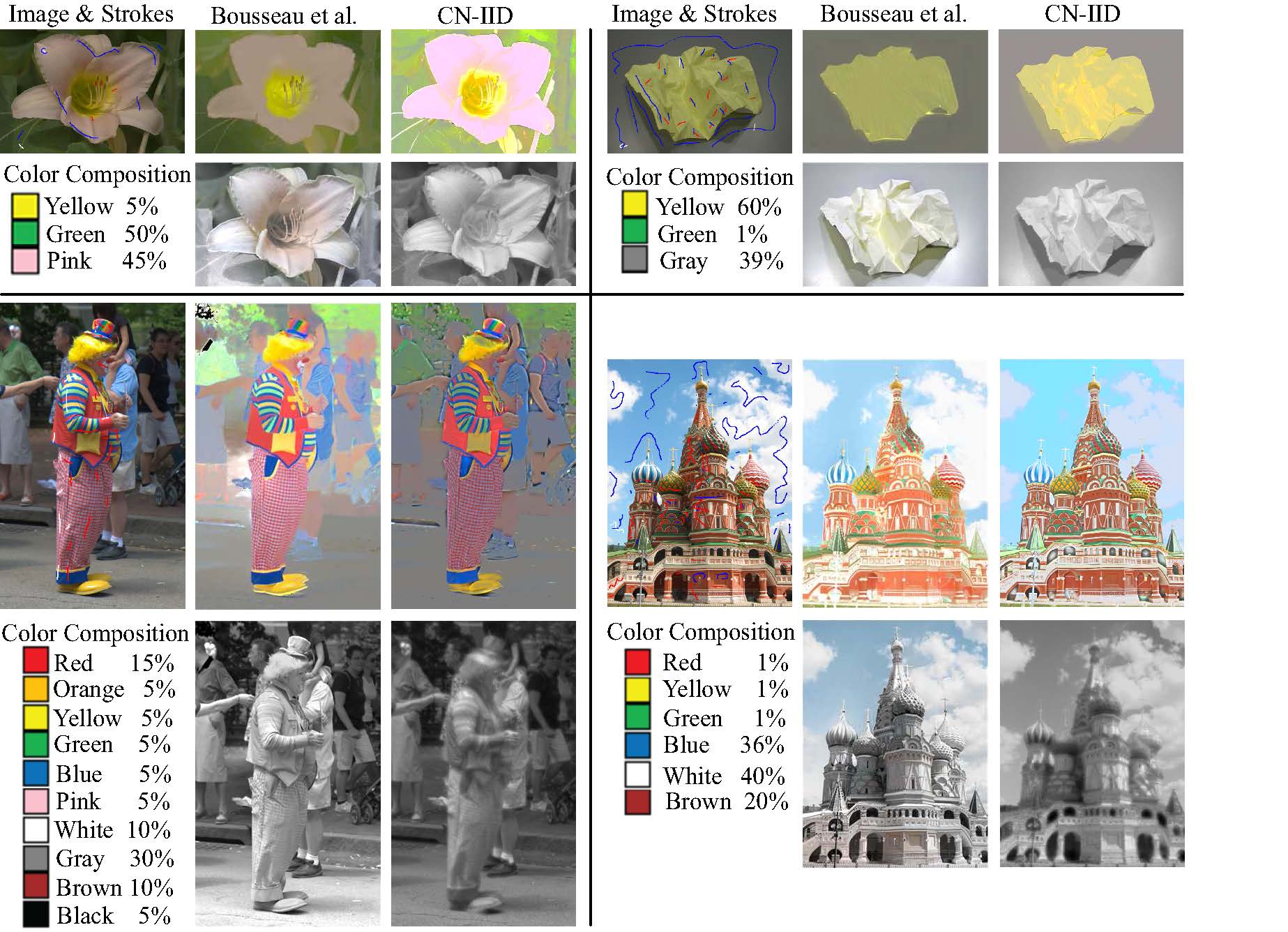}\\
  \caption{Comparison to scribble-based user interaction. The user scribbles are used by the method of \cite{BousseauTOG2009}. White scribbles indicate fully-lit pixels. Blue scribbles correspond to pixels sharing a similar reflectance. Red scribbles correspond to pixels sharing a similar illumination. The color compositions are used by CN-IID.}\label{fig:CN_IID_compare2strokes}
\end{figure*}

The traditional way of user interaction is proposed by \cite{BousseauTOG2009}. They draw a lot of scribbles to tell which areas have constant reflectance or shading. Typically, dozens of strokes are required for obtaining good results, as shown in Fig. \ref{fig:CN_IID_compare2strokes}. Drawing these scribbles is quite time consuming. In comparison, CN-IID needs to input only eleven numbers (the percentages of basic color terms), and the results are comparable to the results of \cite{BousseauTOG2009}. Since humans are quite familiar with basic color terms, they can give a rough annotation quickly, often within a few seconds. For complex images such as those in the bottom row of Fig. \ref{fig:CN_IID_compare2strokes}, the precise color compositions are unavailable. Nevertheless, rough color compositions work well enough in these cases.

\begin{figure*}
  \centering
  \includegraphics[width=1.0\linewidth]{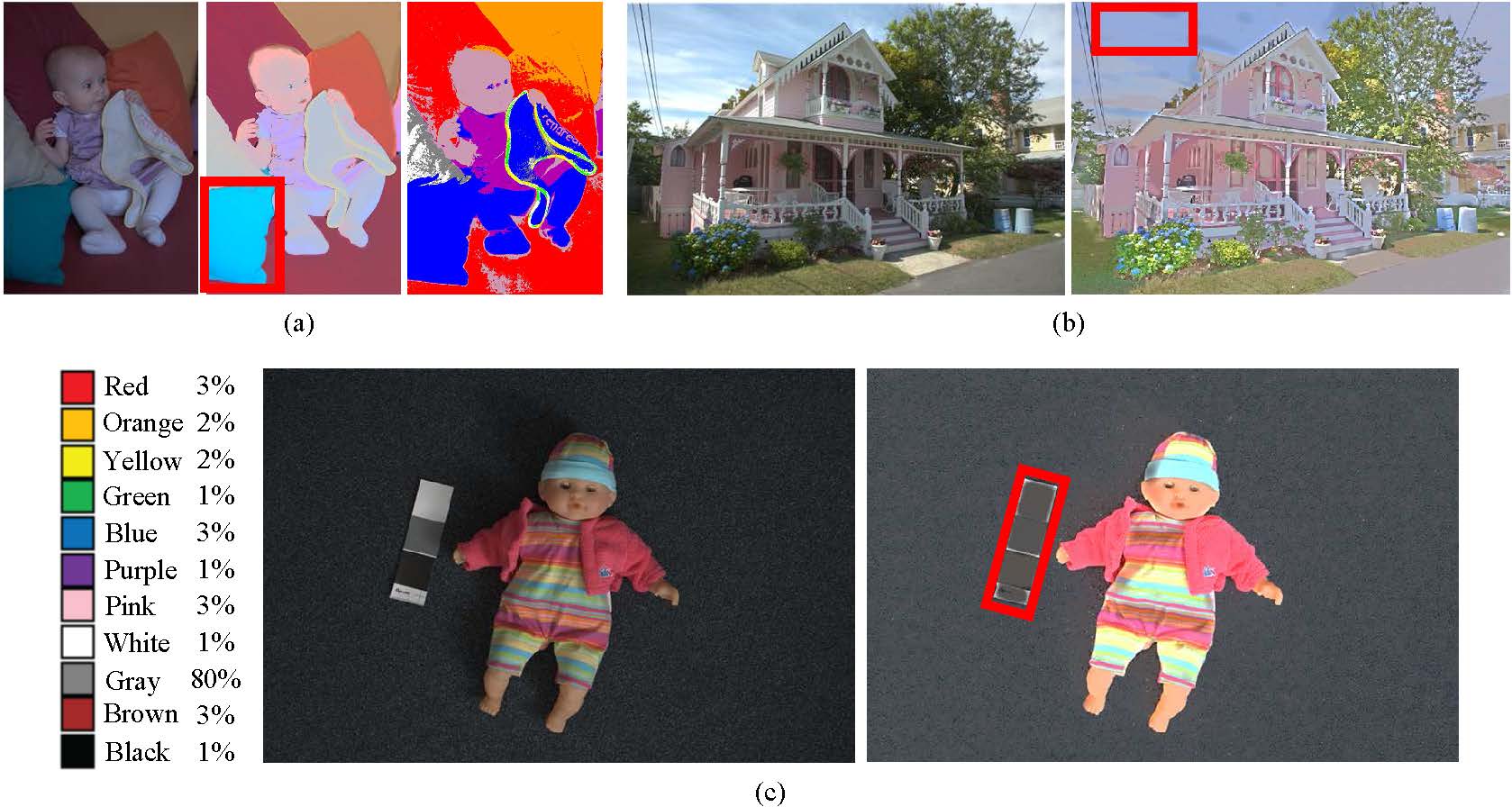}\\
  \caption{Failure cases. The red boxes indicate typical regions where CN-IID failed to recover the reflectance. (a) Intra-category variance of color names makes the recovered reflectance deviate from the perceived color. (b) Texture are erased by cluster-based method. (c) Minor categories are sacrificed.}\label{fig:CN_IID_failure_cases}
\end{figure*}

Image-level color composition cannot determine the reflectance precisely, however. Several typical failure cases are shown in Fig. \ref{fig:CN_IID_failure_cases}. One important defect of CN-IID is the intra-category variation of color names: Each color category covers a large area of the color space \cite{Serra_CVPR12}, so the precision of the color of the recovered reflectance is not high enough. For example, in Fig. \ref{fig:CN_IID_failure_cases}(a), the pillow in the red box appears blue in the recovered reflectance, but not exactly the same blue that we perceived from the image. Another flaw is the block effect that can be observed in the sky of Fig. \ref{fig:CN_IID_failure_cases}(b). Textures and smooth change of color are removed when the reflectance of each cluster is unified. This property is shared by most cluster-based methods \cite{Bell_Siggraph14,chang14svdpgmm,gehler11nips,Garces2012,ShenCVPR08,Serra_CVPR12}. Moreover, the minor categories, \eg white and black in Fig. \ref{fig:CN_IID_failure_cases}(c), may possibly be merged into other categories.

\section{Conclusions and Discussions}

We have presented a new method of user-assisted intrinsic image decomposition that takes color composition as input. The color composition afforded informative constraints on the overall distribution of reflectance, which made intrinsic image decomposition more solvable. In our experiments the performance was improved considerably over the baseline, even when only a rough color composition was available. Compared to the traditional scribble-based user interaction, color naming is efficient yet effective.

We built a generative model to combine the top-down guidance and the bottom-up evidence together. These two sources of information were fused in a probabilistic framework, which give the best explanation of both user perception and image features. However, the features we used here are quite simple, which may not be discriminative enough for all kinds of scenes. Utilizing more powerful features, such as those learned from deep learning \cite{IID_dl}, might further improve the results.

The image-level color composition has a low resolution, so it cannot determine the reflectance precisely. To overcome this limitation, we consider to incorporate the scribbles into CN-IID in our future work. In this way both the global distribution and the details of intrinsic images can be recovered precisely. It is expected that only a few scribbles are needed when the global structure has already been specified by color composition.

Illumination-robust color naming is a challenging task \cite{Color_Naming}. Humans may also make mistakes in this task, especially when they are misled by color illusions. In this case color guidance may distract intrinsic image decomposition. We should better discard the color naming guidance if we are not sure about what the real colors are.

%

%
%
%
\section*{Acknowledgment}
This work was supported by National Basic Research Program of China (973 Program) [No. 2015CB351703] and [No. 2012CB316400]; and 111 Project [No.B13043].

\bibliographystyle{IEEEtran}
\bibliography{refs}

\end{document}